\documentclass[letterpaper, 10 pt, conference]{ieeeconf}  

\IEEEoverridecommandlockouts                              

\usepackage{xcolor}
\usepackage{array}
\usepackage{url}
\usepackage{graphicx}
\usepackage{amsmath}
\usepackage{subcaption}
\usepackage{adjustbox}
\usepackage{makecell}
\usepackage{booktabs}
\usepackage{tabularx}
\usepackage{blindtext}
\let\labelindent\relax
\usepackage{enumitem} 
\setlist{nolistsep} 
\setlist{leftmargin=3mm}

\title{\LARGE \bf
Ethically-Aware Participatory Design of a Productivity Social Robot for College Students
}

\author{Himanshi Lalwani$^{1}$ and Hanan Salam$^{1}$
\thanks{$^{1}$SMART Lab, Department of Computer Science, New York University, Abu Dhabi
        {\tt\small {hanan.salam@nyu.edu, himanshi.lalwani@nyu.edu}}}%
}

\begin{document}

\maketitle
\thispagestyle{empty}
\pagestyle{empty}

\begin{abstract}

College students often face academic and life stressors affecting productivity, especially students with Attention Deficit Hyperactivity Disorder (ADHD) who experience executive functioning challenges. Conventional productivity tools typically demand sustained self-discipline and consistent use, which many students struggle with, leading to disruptive app-switching behaviors.
Socially Assistive Robots (SARs), known for their intuitive and interactive nature, offer promising potential to support productivity in academic environments, having been successfully utilized in domains like education, cognitive development, and mental health. To leverage SARs effectively in addressing student productivity, this study employed a Participatory Design (PD) approach, directly involving college students and a Student Success and Well-Being Coach in the design process. Through interviews and a collaborative workshop, we gathered detailed insights on productivity challenges and identified desirable features for a productivity-focused SAR. Importantly, ethical considerations were integrated from the onset, facilitating responsible and user-aligned design choices.
Our contributions include comprehensive insights into student productivity challenges, SAR design preferences, and actionable recommendations for effective robot characteristics. Additionally, we present stakeholder-derived ethical guidelines to inform responsible future implementations of productivity-focused SARs in higher education.

\end{abstract}

\section{INTRODUCTION}
College students face many academic stressors, including adapting to a new environment, managing demanding coursework, balancing time effectively, navigating classroom competition, coping with fear of failure, and meeting family expectations \cite{akhtar2024issues, freire2020coping, liu2019changes}. These challenges not only hinder productivity, but also contribute to anxiety and depression, ultimately reducing psychological well-being \cite{paul2024crisis}. In fact, research indicates that undergraduate college students experience depression at a higher rate than the general population \cite{auerbach2016mental}. 
The issue becomes more pronounced amongst students diagnosed with Attention
Deficit Hyperactivity Disorder (ADHD) as they experience heightened difficulties with executive functioning \cite{weyandt2008adhd, heiligenstein1999psychological}. 
Although productivity tools like planners and mobile apps exist, their effectiveness depends on self-discipline and consistent use, which many students find difficult to sustain. Moreover, they also cause constant app switching which disrupts focus \cite{qatalog2021workgeist}. Hence, there is a need for more effective and engaging support systems that help students develop sustainable productivity habits.

Socially Assistive Robots (SARs) \cite{feil2005defining} offer a promising solution to college students' productivity challenges by providing natural, intuitive interactions. SARs have been successfully used across various domains, including promoting creativity in children \cite{elgarf2024fostering}, teaching support in kindergartens \cite{fridin2014storytelling}, social development for children with autism \cite{wood2021developing}, and cognitive therapy for older adults \cite{lozano2023open}. These applications highlight SARs' potential as innovative, personalized tools for addressing academic productivity pressures.

However, to fully realize the potential of social robots in supporting college students' productivity, it is essential that these systems are designed with direct input from the students themselves. Participatory Design (PD) is a collaborative design approach that involves multiple stakeholders in the design process to ensure that the resulting product addresses their needs, challenges, and expectations \cite{bjogvinsson2012design}. Previous studies have used PD to develop SARs in mental health. For example, PD was used to design a SAR  for older adults with depression in \cite{vsabanovic2015robot}, while~\cite{moharana2019robots} collaborated with dementia caregivers to conceptualize a SAR for caregiving support. In \cite{rose2017designing}, the authors engaged middle and high school students to explore the design of a social robot for measuring teen stress, and in~\cite{9515356}  insights were gathered from users and experienced coaches to inform the development of a robotic well-being coach. These studies highlight the value of PD in creating user-centered solutions. 

In this study, we applied Participatory Design to actively involve college students and a Student Success and Well-Being Coach in the design of a SAR for productivity, ensuring that the resulting productivity-focused bot aligns with their needs and preferences. To achieve this, we conducted interviews with college students to explore their productivity challenges and strategies, followed by a group workshop where selected students collaborated to brainstorm the robot’s design. Additionally, we interviewed a Student Success and Well-Being Coach at the local institution to gather similar insights. Unlike existing PD studies, we prioritized ethics from the outset by engaging participants in discussions about ethical considerations, ensuring that their perspectives shaped both the design and responsible deployment of the productivity-focused bot. 
\section{RELATED WORK}


\paragraph{Socially Assistive Robots for College Students}
SARs have found applications in various domains to support college students.  In~\cite{rice2023effectiveness}, Pepper robot was used  to study the effectiveness of social robots in facilitating mental well-being interventions such as stress-reducing exercises among university students. Similarly, in~\cite{9223588}, a robotic positive psychology coach aimed at enhancing the psychological well-being of college students was developed. In another study, a Socially Interactive Robotic Tutor that provides verbal encouragement to increase student engagement in mathematics education was created \cite{6891009}. Alex, a robotic companion that assists college students in time management and focused work sessions, was proposed in \cite{10974238}. Lastly,~\cite{O'connell} developed Blossom, a minimally interactive in-dorm robotic study companion, that supported college students with ADHD during their work. While these solutions have shown promising results, none of these employed a participatory design approach with the students. By involving students in the co-design process, this study presents multiple productivity-focused robot prototypes developed by the students based on their needs and perspectives. The insights gained provide a foundation for developing SARs tailored to academic settings.

\paragraph{Attention Deficit Hyperactivity Disorder}
ADHD is a chronic neurodevelopmental condition marked by significant inattention, hyperactivity, and impulsivity \cite{furman2005attention}. While typically diagnosed during childhood, its symptoms persist into adolescence and adulthood, affecting an estimated 2–8\% of college students \cite{Resnick_2005}. Students with ADHD often face compounded challenges—ranging from lower academic performance and reduced self-esteem to heightened risks of depression, anxiety, and social isolation—which can severely compromise their productivity and overall well-being \cite{DuPaul_Weyandt}. Given the extensive presence of ADHD among college students and its impact on productivity, it becomes important to involve students with ADHD in our co-design process. Moreover, the scarcity of research on SARs for young adults with ADHD highlights the critical value of their insights in shaping effective solutions. At the same time, incorporating neurotypical students provides complementary perspectives, ensuring that our designs are both adaptable and inclusive across diverse cognitive profiles.

\paragraph{Ethics by Design}
The inclusion of ethical considerations in participatory design processes is increasingly critical in social robotics. 
Recent literature underscores ethical challenges beyond privacy, including fairness, transparency, and accountability  \cite{brey2024ethics,ostrowski2022ethics}. Ostrowski \textit{et al.} \cite{ostrowski2022ethics} specifically highlight the limited engagement with ethics, equity, and justice in HRI research, advocating for explicit incorporation of these considerations into robotic design to address systemic biases and promote equitable interactions.
\cite{hitron2023implications} show that unnoticed AI biases in robotic behavior can reinforce harmful stereotypes. Their findings illustrate the importance of embedding explicit ethical awareness and unbiased behaviors into robotic systems to mitigate unintended societal impacts.
Moreover, the ``Ethics by Design'' framework  \cite{dignum2018ethics} emphasizes that autonomous systems must inherently possess the capability to reason ethically, remain within moral bounds, and transparently justify their actions. This  work provides a robust rationale for integrating ethics directly into the design phase, thereby facilitating the development of trustworthy  systems. By including questions about ethics in our participatory design, we align our research with these established frameworks and address critical ethical challenges proactively, ensuring our robotic coach effectively serves diverse user needs without unintended ethical implications.

\section{METHODS}
To understand college students' needs for a productivity-supporting robot, we conducted a participatory design including individual interviews and a group workshop with students facing productivity challenges, and a Student Success and Well-Being Coach. The study was approved by the Institutional Review Board (IRB \#HRPP-2024-125).


\subsection{Participants} 
\begin{figure}[ht]
    \centering
    \begin{subfigure}[b]{0.45\columnwidth}
        \centering
         \captionsetup{justification=raggedright, singlelinecheck=false, labelfont=bf, labelsep=period, font=small} 
        \caption{}
        \includegraphics[width=\columnwidth]{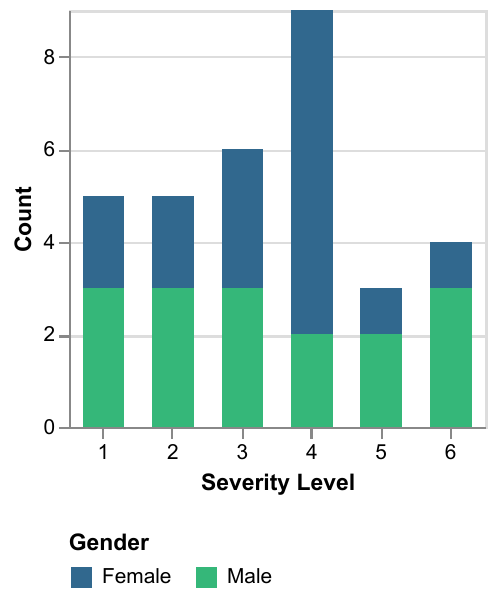} 
        \label{fig:asrs}
    \end{subfigure}
    \hfill
    \begin{subfigure}[b]{0.49\columnwidth}
        \centering
         \captionsetup{justification=raggedright, singlelinecheck=false, labelfont=bf, labelsep=period, font=small} 
        \caption{}
        \includegraphics[width=\columnwidth]{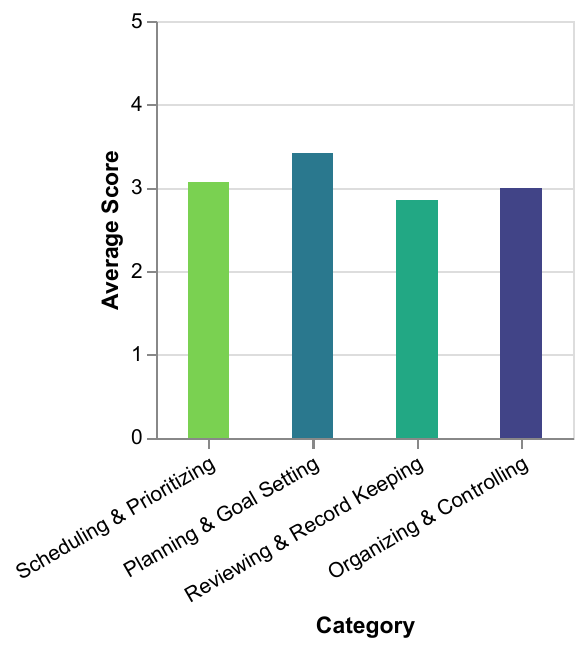} 
        \label{fig:stms}
    \end{subfigure}

    \caption{(A) Distribution of participants based on ASRS scores. (B) Average scores across different STMS management categories.}
    \label{fig:participant-profiles}
\end{figure}

We advertised our study through flyers inviting full-time undergraduate students, aged 18 to 24, who were facing productivity challenges to participate. The flyers included a QR code linking to an online sign-up form, which collected demographic information, self-reported productivity levels on 5-point Likert scale, and the Adult ADHD Self-Report Scale (ASRS) \cite{kessler2005world}. 
A total of 75 students from the local university signed up. Invitations to participate in 1-on-1 interviews (cf. section \ref{sec:individual-interviews}) were sent to all students who rated their productivity at most 3. The final participant pool comprised 32 students (17 freshmen, 7 sophomores, 1 junior, and 7 seniors) from 15 different majors. Participants were evenly split by gender, with 16 identified as ADHD and 16 as non-ADHD based on their ASRS scores (see Figure \ref{fig:asrs}). Each participant was compensated with a gift voucher at the end of the interview. 

10 students, 5 males and 5 females, selected from the 32 who participated in the individual interviews were invited to participate in a group workshop. Students who indicated in the interviews that they did not rely on technology for productivity were excluded, and 10 participants were randomly chosen from the remaining pool, with attention to gender diversity and ASRS scores. 1 participant did not attend the workshop, resulting in a final group of 9 participants. Based on the ASRS scores, 4 participants had ADHD, whereas the remaining 5 did not. Upon completion of the workshop, each participant was compensated with a gift voucher. 

Mental Health and Well-Being Counselors and Student Success and Well-Being Coaches at the university were also invited via email to participate in 1-on-1 interviews, though only one agreed to join the study.

\subsection{Undergraduate Students Interviews}
\label{sec:individual-interviews}
 We conducted 1-on-1 interviews with the recruited undergraduate students to gain insights into their productivity challenges and the tools and strategies they use to enhance productivity in their daily lives. Additionally, we asked them for their thoughts on using a social robot designed to enhance productivity, including any ethical concerns they might have about such a system. Each interview lasted approximately 15 to 20 minutes and was conducted via Zoom and video recorded for analysis. 
Participants completed a pre-interview survey, which included the Student Time Management Scale (STMS) \cite{article} and the Negative Attitudes Toward Robots (NARS) scale \cite{nomura2006experimental} to assess their time management skills and perceptions of robots. STMS results revealed moderate difficulties across all categories (see Figure \ref{fig:stms}).

\subsection{Student Success and Well-Being Coach Interview}
We also interviewed a Student Success and Well-Being Coach from the same institution, who has over 10 years of experience supporting college students with academic challenges. The interview focused on understanding the productivity challenges college students commonly face, learning about some proven strategies the coach recommends, and discussing the potential role of a social robot in supporting productivity. This 35-minute interview was conducted online via Zoom and video recorded for analysis.

\subsection{Group Workshop}
\label{sec:group-workshop}
The workshop lasted approximately three hours and featured interactive, hands-on activities to encourage participants' active input on the development of a productivity-focused robot. The workshop started with a welcome and a brief introduction. It was structured as follows: 

\textit{\textbf{Identifying Common Productivity Challenges.}} It began with a warm-up discussion in which participants revisited their productivity challenges. Each participant was asked to share at least 3 challenges on Mentimeter, an interactive presentation tool. Additionally, they were asked to vote on others' responses they could relate to, helping identify the most common challenges. Then the participants discussed their responses with the rest of the group. This brief review was intended to refocus on their challenges, setting the foundation for the solution-oriented activities that followed. 

\textbf{\textit{Robo Showdown: Exploring Social Robot Designs.}} In the next segment, participants were introduced to different social robots. We showed them videos of Pepper, Paro, Furhat, Jibo, and QTrobot. These robots were chosen to give participants an overview of the different possible embodiments of social robots that have already been developed. After watching, participants provided verbal and written feedback on the features and designs of each robot. At the end, they also ranked the robots based on their preference for the best-suited design for a potential productivity-focused robot. 

\textbf{\textit{Team Challenge.}} In the final segment, participants were paired up for a competitive challenge that involved two activities. There were four teams: Teams A, B, and C had two participants each, and Team D had three participants.
The grouping was designed to explore how team composition influences design outcomes by comparing an ADHD-only team (Team A) with a non-ADHD team (Team B) and mixed teams (Teams C and D) that mirror real-world diversity.

\begin{figure}
  \centering
  \includegraphics[scale=0.25]{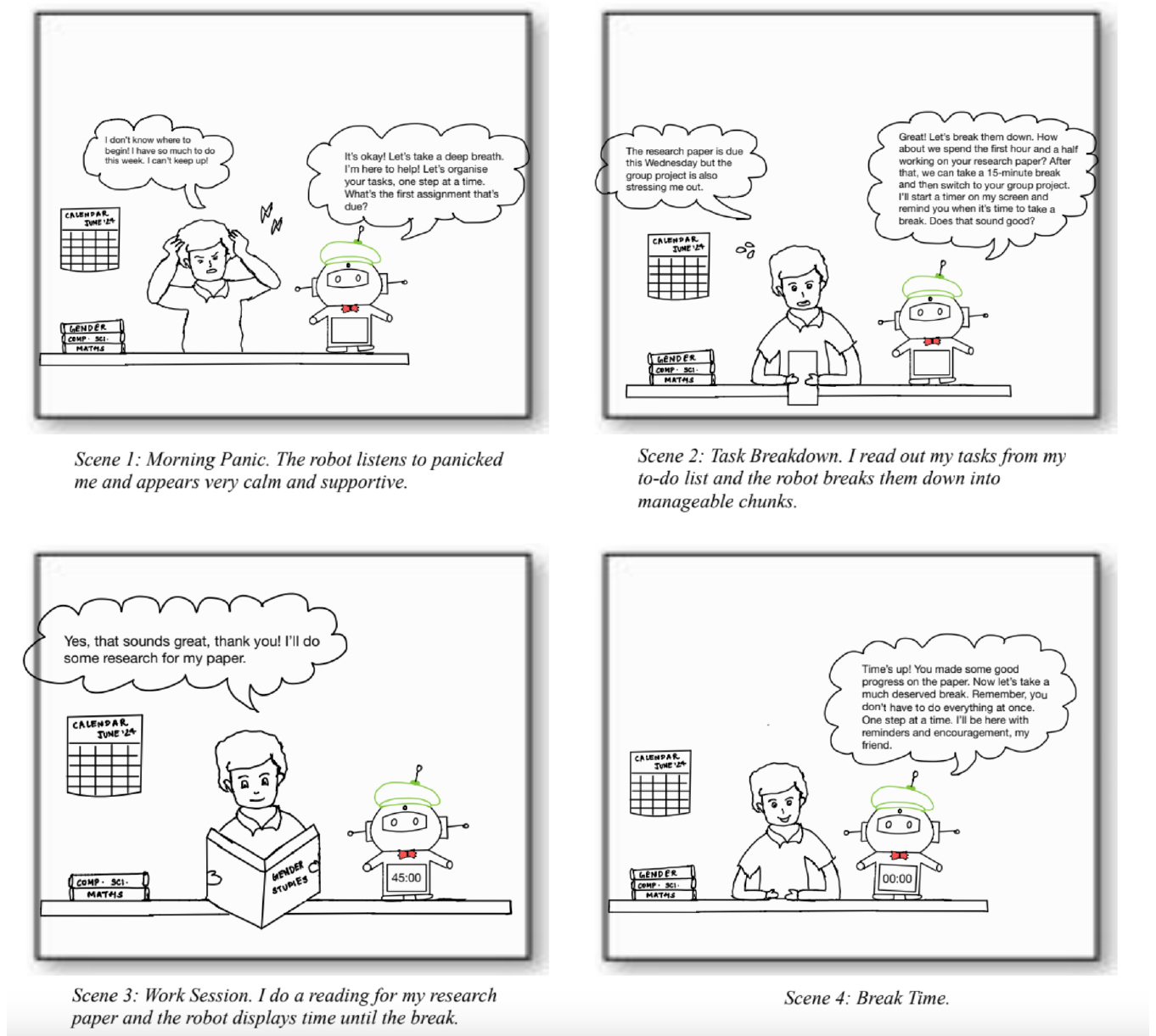}
  \caption{Sample Storyboard shared with Participants.}
  \label{fig:example-stroyboard}
\end{figure}

\textit{Scenario Creation Activity}: Each team identified a personal productivity challenge and created a storyboard showing how a social robot designed for productivity enhancement can help overcome the challenge. We provided them with a template and an example (see Figure \ref{fig:example-stroyboard}) for guidance. They had 20 minutes to complete this activity, after which each team presented their scenario to the group.

\textit{Prototype Design Activity}. Each team created a prototype of their productivity-focused bot and developed a persona for it. We used Miro, an online whiteboard platform, for this activity. Each team was given a template with space for prototyping, guiding prompts for persona creation, and questions encouraging reflection on the ethical considerations involved in designing their prototype. Participants had the option to choose between creating a digital or physical prototype. They had 30 minutes to complete this task, followed by presentations to the group.

After both activities, each team ranked other team's storyboards and prototypes using a Google Form. The principal investigator of this study also assessed the work. The team with the highest aggregate score won a gift voucher, in addition to the participation incentive.

At the end of the workshop, participants completed a post-workshop questionnaire which included a second administration of the NARS scale to measure changes in their attitudes towards robots after their introduction in the workshop. It also asked general questions about a productivity-focused robot, such as its most beneficial features and whether participants would recommend it to others. 





\subsection{Data Analysis Method}
We used thematic and content analysis to analyse the qualitative data. Thematic analysis helps identify, analyse, and interpret patterns of meaning or themes within qualitative data \cite{boyatzis1998transforming}. Content analysis is used to quantify and analyse the presence, meaning, and relationship of such themes \cite{krippendorff2018content}. We selected thematic analysis to draw out themes from the interviews and group workshop, and content analysis to quantify their frequency and rank them by significance.

\section{FINDINGS}

\subsection{Quantitative Findings}
Participants' attitudes toward robots remained unchanged after the workshop, as measured by NARS ($p = 0.2$). Participants in the workshop rated their likelihood of recommending a productivity-focused robot at an average of 4, while their likelihood of using it was 3.7 on a 5-point scale, suggesting a generally positive reception of the concept.

\subsection{Qualitative Findings}
\subsubsection{Common Productivity Related Challenges}
The following challenges, listed in order of significance, were most frequently reported by participants during both individual interviews and the group workshop:

\begin{itemize} \item \textbf{Procrastination.} Students often delay tasks until the last minute, spending time on social media or engaging in other non-urgent activities. For example, P8 mentioned that they would clean their room or focus on another task that did not have an immediate deadline. P6 and P10 shared similar experiences, explaining that they feel they have enough time in the moment and they need the pressure of impending deadlines to get into a productive mindset.
\item \textbf{Lack of Motivation.} Students also struggle with a lack of motivation to begin tasks. Participants mostly mentioned doing tasks out of obligation rather than interest. However, several sources of extrinsic motivation were identified. These included deadlines, stress, and past experiences of failing to complete tasks on time and the resulting negative impact on their work quality. Additionally, the relief felt after completing a task and the fear of failure, especially the consequences on GPA, grad school admissions, and future plans, were considered strong motivating factors. For some students, the sight of peers working around them also served as a reminder to stay productive. P1 shared that being far from home, halfway across the world, motivates them to remain focused. They remind themselves that they are in college to learn and gain experiences, and wasting time would detract from this opportunity.
\item \textbf{Time Management and Task Prioritization.} Balancing multiple commitments leads students to sometimes prioritize non-urgent or less important activities. For example, P12 and P21 shared that, despite creating daily schedules, they struggle to stick to them due to spending extra time with friends driven by the fear of missing out or dealing with unexpected events at the university. They prioritized these activities over their tasks. The Student Success and Well-Being Coach also highlighted this issue, adding that due to poor time management, students often feel they spend more time on uninteresting tasks than they actually do and struggle to assess their performance on these tasks in relation to the time spent. Participants mentioned that actively reflecting on task significance, breaking tasks into smaller chunks, and incorporating flexible time blocks into their schedules helped manage these challenges. 
\item \textbf{Task Characteristics and Interest Levels.} Productivity on a task is influenced by how engaging and complex students perceive the task to be. Tasks that are boring, large, or difficult often lead to procrastination. Conversely, when students find the subject matter interesting or manageable, they are more likely to complete it with ease. Hence, students often try to connect tasks to something of personal interest to make them more engaging. Additionally, many participants mentioned rewarding themselves after completing boring or difficult tasks as a way to stay motivated.
\item \textbf{Focus and Attention.} Sustaining focus for more than 30 minutes is challenging, especially when tasks are demanding. As a result, students often switch between tasks without completing any of them effectively. To minimize this, many have adopted strategies such as taking regular breaks and using Pomodoro timers to maintain concentration. \end{itemize}

\subsubsection{Tools for Productivity Enhancement} 
During individual interviews, participants mentioned various digital and physical tools they use to enhance productivity and manage cognitive load. \textit{Google Calendar} was most popular due to its user-friendly interface and customizable features like tailored reminders, detailed event entries, and color-coding for different categories, though some noted issues like easily overlooked notifications and visual clutter.  
\textit{Notion} was frequently mentioned, though opinions were mixed. Some praised it for centralized note-taking, task management, and customizable pages for organizing class information. Others found it overwhelming due to numerous layouts and customization features, which sometimes became distractions. The absence of built-in task reminders was also a notable drawback.
Those who struggled with Notion preferred simpler alternatives like the iPhone’s built-in \textit{Notes app} for creating to-do lists, scheduling, and journaling. They valued its minimalist design, ease of use, quick access, and the satisfaction of checking off completed tasks.
 Many relied on the \textit{Reminders app} for additional alerts alongside Google Calendar. Less commonly used tools included \textit{Flora} (Pomodoro timers), \textit{Opal} (app blocking), \textit{Microsoft Task Manager}, \textit{Google Keep}, and \textit{ChatGPT} (for idea generation). A minority favored \textit{physical tools}, like notebooks and whiteboards, to minimize digital distractions and experience the tangible satisfaction of marking tasks completed.
The insights into each application's strengths and weaknesses provide valuable guidance for designing a robotic productivity system. Such a system must combine the best features of existing tools to address the full spectrum of individual cognitive and work style needs.

\subsubsection{Robot Form}
\begin{figure}
  \centering
  \includegraphics[scale=0.19]{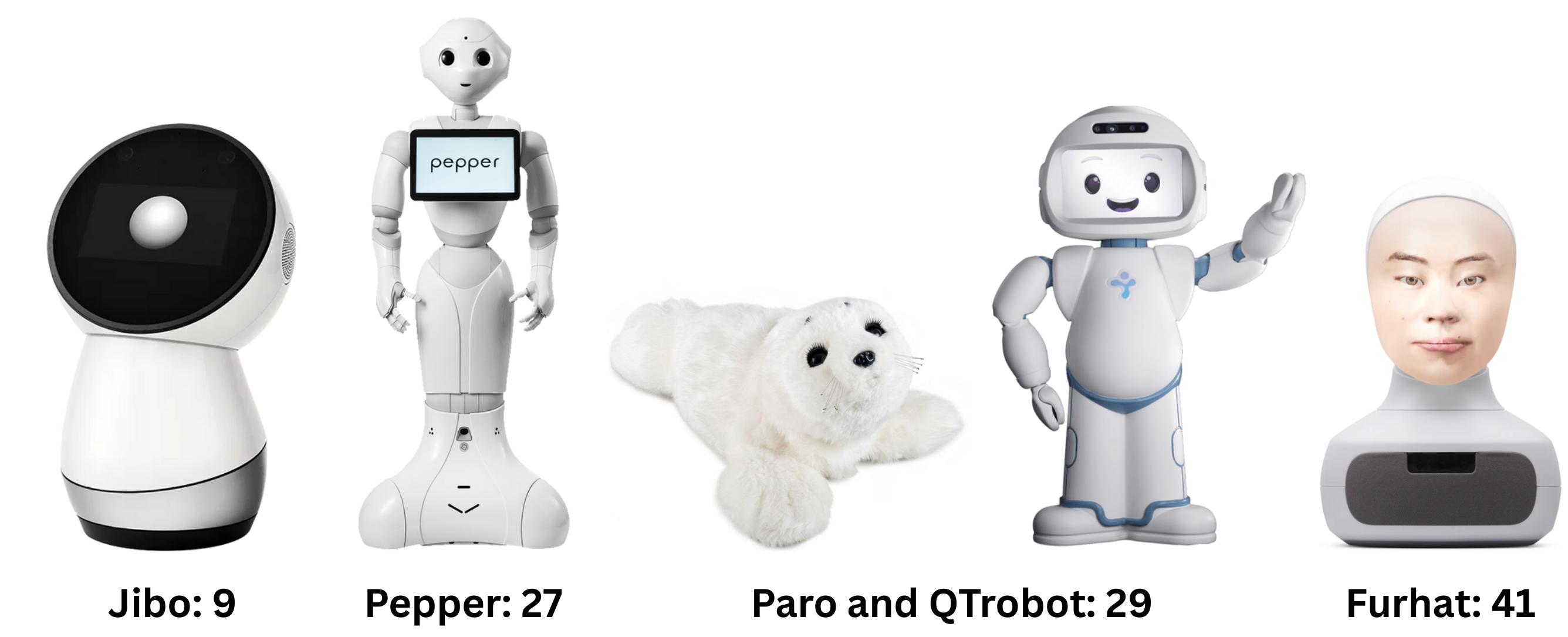}
  \caption{Aggregate preference scores for the five robots; lower scores reflect higher overall preference.}
  \label{fig:image3}
\end{figure}

In the group workshop, participants ranked 5 robots (Furhat, Jibo, Paro, Pepper, and QTrobot) from most (1) to least (5) preferred form for a productivity-focused robot. Jibo received the lowest total score, indicating the highest preference, followed by Pepper, with Paro and QTrobot tied. Furhat was the least preferred form. The scores for each robot are shown in Figure \ref{fig:image3}.
Participants preferred a small, abstract, and portable robot with a screen, that could be easily carried and used anywhere without causing disturbance to people nearby. They also disliked a childlike appearance in robots, feeling it would make them less inclined to take the robot’s guidance seriously. Additionally, they preferred the robot not to have an overly human-like face as this would make it unsettling to use.

\subsubsection{Robot Prototypes}
\begin{figure}[ht]
    \centering
    \begin{subfigure}[b]{0.49\columnwidth}
        \centering
         \captionsetup{justification=raggedright, singlelinecheck=false, labelfont=bf, labelsep=period, font=small} 
        \caption{}
        \includegraphics[width=\textwidth]{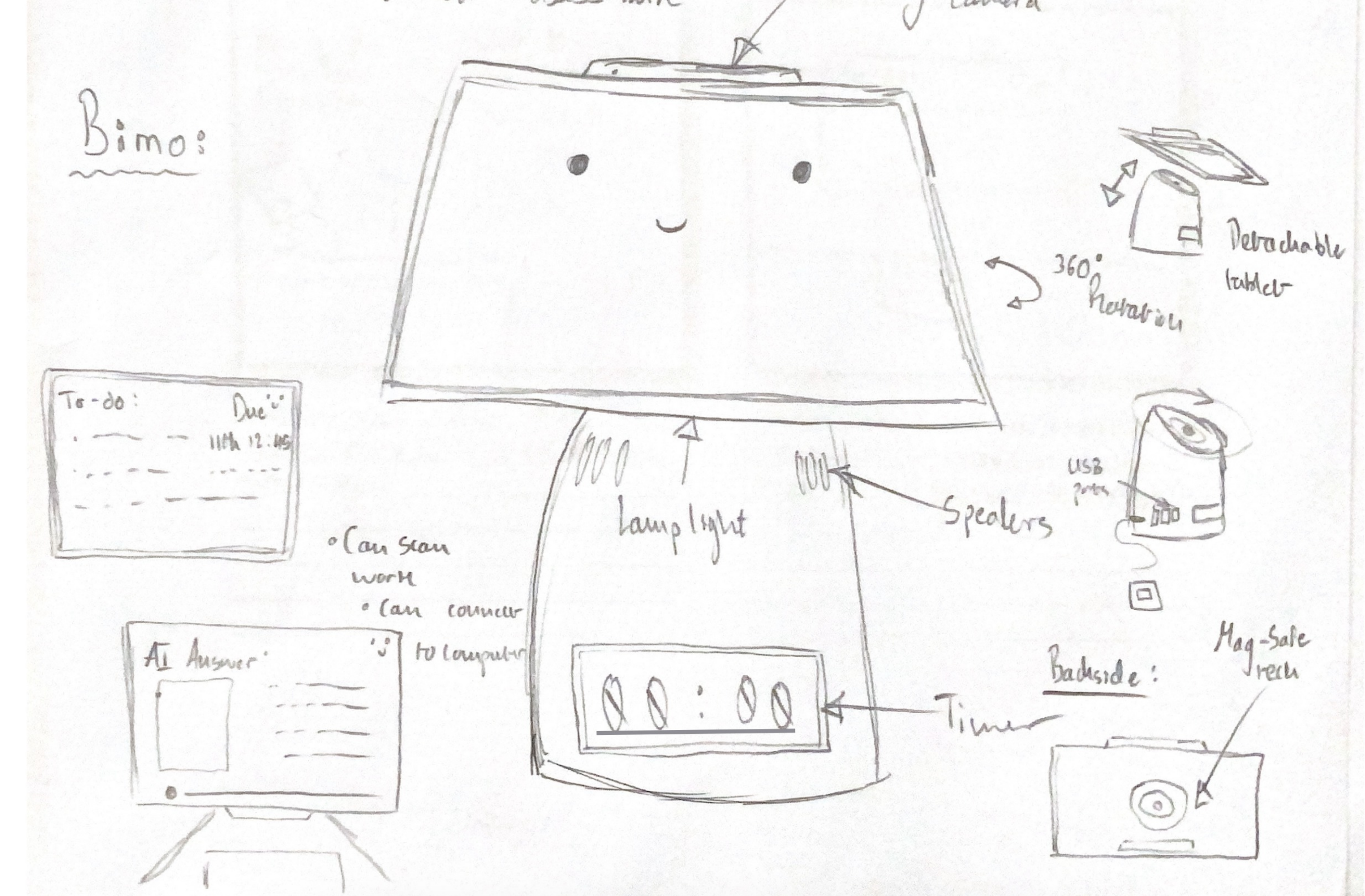} 
        \label{fig:Bimo}
    \end{subfigure}
    \hfill
    \begin{subfigure}[b]{0.49\columnwidth}
        \centering
         \captionsetup{justification=raggedright, singlelinecheck=false, labelfont=bf, labelsep=period, font=small} 
        \caption{}
        \includegraphics[width=\textwidth]{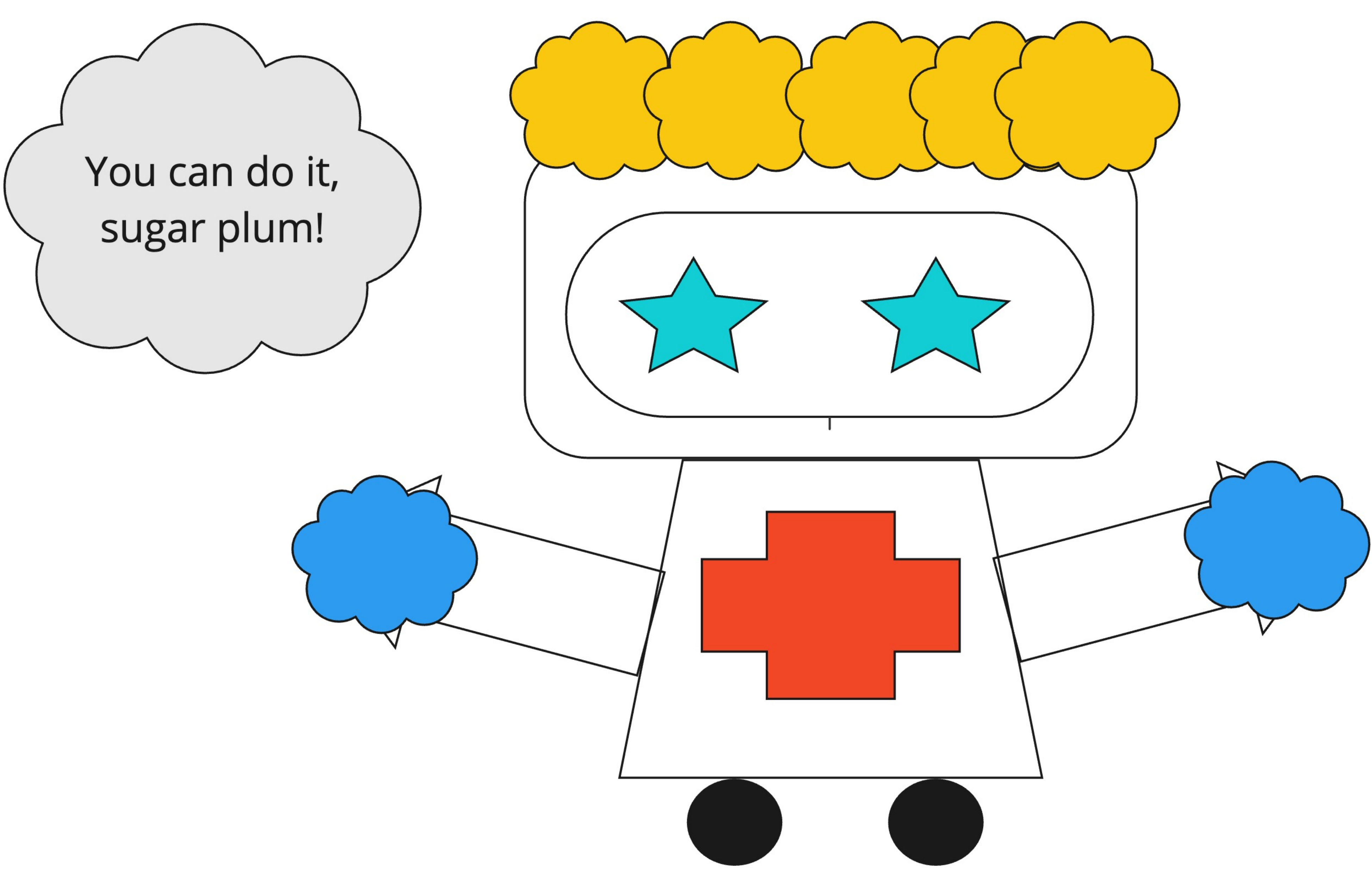} 
        \label{fig:Max}
    \end{subfigure}

    \vspace{1em} 

    \begin{subfigure}[b]{0.49\columnwidth}
        \centering
         \captionsetup{justification=raggedright, singlelinecheck=false, labelfont=bf, labelsep=period, font=small} 
        \caption{}
        \includegraphics[width=\textwidth]{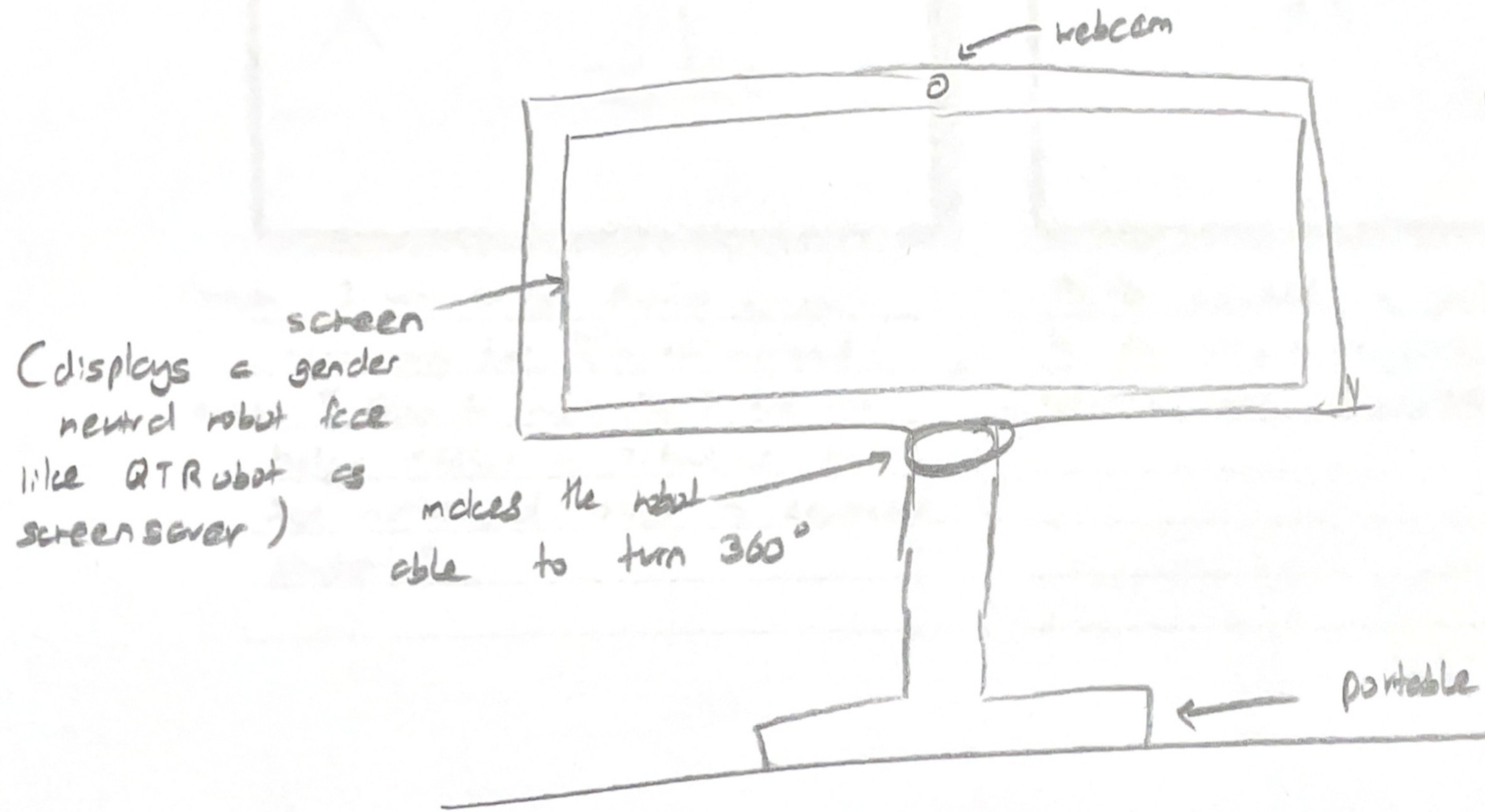} 
        \label{fig:Roberto}
    \end{subfigure}
    \hfill
    \begin{subfigure}[b]{0.49\columnwidth}
        \centering
         \captionsetup{justification=raggedright, singlelinecheck=false, labelfont=bf, labelsep=period, font=small} 
        \caption{}
        \includegraphics[width=\textwidth]{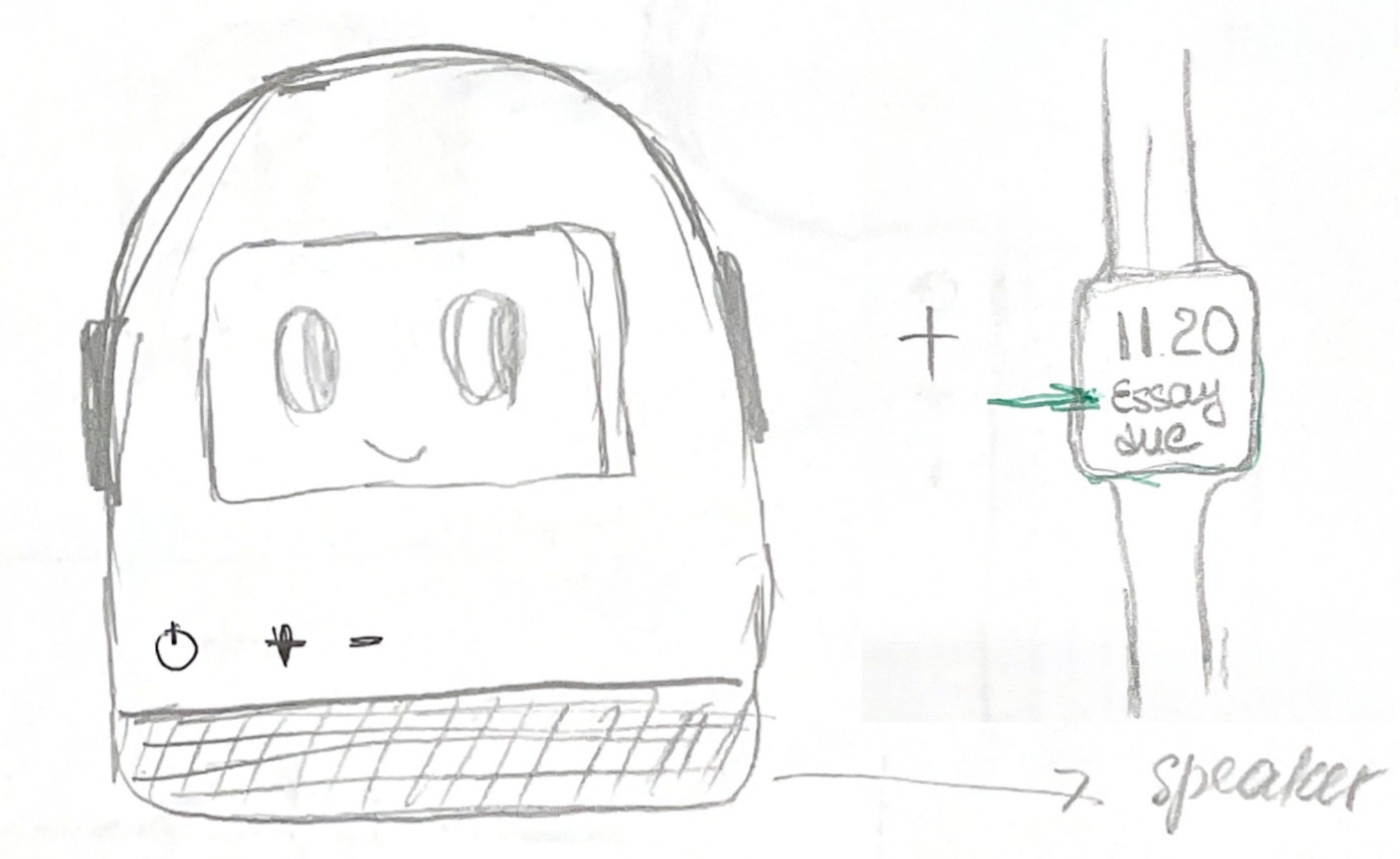} 
        \label{fig:Bammy}
    \end{subfigure}

    \caption{Prototypes Created by Different Teams: (A) Bimo, (B) Max, (C) Roberto, (D) Bammy.}
    \label{fig:prototypes}
\end{figure}

The prototypes developed by the teams during the group workshop are shown in Figure \ref{fig:prototypes}, with further details provided in Table \ref{table:robot_features}.
\begin{itemize}
    \item \textbf{Bimo:} Team A developed Bimo, a gender-neutral robot named after BMO, a character from Adventure Time. Bimo serves as a ``smart friend who can joke around and have fun but can also mentor you when needed.'' It is portable, equipped with a tablet and a lamp that doubles as a speaker.

    \item \textbf{Max:} Team B created Max, a gender-neutral robot that blends human and animal traits. Max functions as an ``assistant (divides tasks), friend (rewards you, plays music), mentor (guiding you, making decisions).'' It features a tablet, wheels for movement, arms that display gestures such as hugging or giving a high-five, and paws that provide warmth for comfort through physical touch.

    \item \textbf{Roberto:} Team C designed Roberto, another gender-neutral robot. Roberto acts as a coach as it ``encourages students to do work, monitors distractions, provides motivation similar to how a coach would.'' It is portable and features a screen that rotates 360 degrees.

    \item \textbf{Bammy:} Team D developed Bammy, a portable robot that acts as both mentor and motivator. The team emphasized that
they did not want a robot that was ``too strict, controlling, and too friendly / too chatty.'' Bammy includes a tablet and smartwatch to monitor stress and health.
\end{itemize}

\begin{table*}[ht]
\centering
\scriptsize
\setlength{\tabcolsep}{5pt}  
\adjustbox{max width=\textwidth}{ 
\begin{tabular}{ m{1.4cm}  m{1.5cm} m{1.1cm} m{2cm}  m{4.5cm} m{4.5cm}}
\toprule
 \textbf{Design (Name/Gender)} & \textbf{Personality} & \textbf{Role} & \makecell{\textbf{Interaction} \\\textbf{Modalities}}   & \textbf{Features} & \textbf{Ethical Considerations} \\
\midrule
\centering Bimo (Gender-Neutral)  &  \makecell[l]{Friendly\\Gentle\\ Firm\\Reactive} & \makecell[l]{Smart Friend\\ Mentor} & \makecell[l]{Speech \\Light Indicators \\Visually via Tablet } & Warm lamp light with built-in speaker; Tablet for content display; rotatable head; Integrates with student's work ecosystem; Monitors screen for careless mistakes; AI-powered feedback on student work; Helps organize and prioritize tasks; Monitors stress and exhaustion via camera & Saves  videos to a student-accessible SD card; Deletes monitored videos after 2 days; Requires manual on/off for student control; Always asks permission before accessing third-party applications; Detects critical health issues (e.g., heart attacks) and alerts authorities promptly \\
\midrule
\centering Max (Gender-Neutral) & \makecell[l]{Friendly\\ Assertive\\ Motivating\\ Strict with\\ deadlines} & \makecell[l]{Assistant\\ Friend\\ Mentor\\Proactive} & \makecell[l]{Speech \\Light Indicators \\Gestures \\Facial Expressions \\Physical Touch}   & Plays study music; Offers effective study tips; Helps with task organization; Comes with wheels for easy portability & Requires manual on/off for student control; Does not store user data; Respects the laws and customs of the region \\
\midrule
\centering Roberto (Gender-Neutral) & \makecell[l]{Friendly\\ Funny\\ Kind\\ Calm\\ Encouraging\\Proactive} & Coach & \makecell[l]{Speech \\Visually via Screen\\ Vibrations }  & Screen for content display; Helps with task organization and prioritization; Monitors for distractions via camera; Offers task feedback; Uses alarms and vibrations to dis-incentivize students from getting distracted & Does not store any monitoring videos \\
\midrule
\centering Bammy (Customizable) &  \makecell[l]{Friendly\\ Organized\\ Nice\\ Straightforward\\ Honest\\Reactive \& Proactive} & Mentor Motivator & \makecell[l]{Speech\\ Visually via screen\\ Visually via watch } & Screen for content display; AI-powered for task organization; Integrates with student’s workflow; Dedicated app; Monitors stress and health via smartwatch; Provides emotional support; Remembers user preferences; Accesses quality research sources; Supports language flexibility & No camera for privacy; Identifies users to protect others' personal information; Accessibility features for individuals with special needs\\
\bottomrule
\end{tabular}
} 
\caption{Key features of prototypes created during the group workshop. The robots varied in their roles and abilities. Reactive (Bimo waits for the student to interact first). Proactive (Max leads the conversations and Roberto constantly checks in with the student)}
\label{table:robot_features}
\end{table*}

\subsubsection{Robot Capabilities}
The robots designs shared several features identified through thematic and content analysis. Here, we discuss these themes in order of significance.
\begin{itemize}
    \item \textbf{Assisting with Managing Tasks and Creating Personalized Schedules.} 
    All prototypes were designed with features to help create schedules based on users' preferred time intervals (weekly, daily, or another interval). They supported task prioritization and  breakdown. Additionally, Bammy featured the ability to personalize schedules according to individual work styles and preferences.
    \item \textbf{Using Timers for Work Sessions and Breaks.} To structure work periods and ensure regular breaks, all teams equipped their robots with visual timers that would help them maintain focus and keep a healthy work-rest balance. 
    \item \textbf{Displaying Content on Screen.} All teams incorporated screens into their robots to display content in the form of educational videos, to-do lists, visualized schedules, deadline notifications, and the robot’s emotions. Roberto further featured the ability to stream entertainment content during breaks, with automatic termination of content when the break ended to prevent extended downtime.
    \item \textbf{Monitoring and Intervention.} Many teams wanted their robots to monitor well-being and focus. Bimo and Roberto included cameras to detect signs of stress, while Bammy relied on health data from a linked smartwatch, addressing privacy concerns with the camera. If stress was detected, the robots could intervene with affirmations or guide the user through mindfulness exercises. In cases of distraction, they could prompt the user to refocus in different ways. For example, Roberto was designed to start with a gentle warning and, if distraction persisted, trigger a vibrating alarm to discourage prolonged disengagement.
    \item \textbf{Offering Rewards.} Since students often struggle with motivation, Bimo and Max were designed to encourage them through rewards. These rewards could include small treats like candies, break time, or emotional encouragement as a form of positive reinforcement.
    \item \textbf{Integrating with the Student's Work Ecosystem.} Students often use a variety of productivity applications in their daily lives. To address this, some teams envisioned a robot that could unify these digital tools. Bimo and Bammy featured dedicated applications that could connect to learning management systems, calendars, and to-do lists, helping students avoid the hassle of switching between apps and streamline their workflow.
    \item \textbf{Providing Task Feedback.} Some teams suggested that the robot could provide constructive feedback on their tasks. For example, Bimo was designed to allow students to share assignment grading rubrics, which it could then use with generative AI to generate comments aligned with the rubric. Since students often rush through assignments, Bimo could also monitor screens to detect careless errors and ensure work quality. Additionally, Roberto could help students prepare for presentations by listening to their rehearsals and offering feedback using similar technology.

\end{itemize}

\subsubsection{Interaction with the Robot}
Participants proposed various interaction modalities for their prototypes. Speech emerged as the most essential, with participants wanting the robot to have a natural voice, one that sounded less robotic and more engaging. Visual communication was also prioritized, with participants envisioning the robot using its screen for reminders, timers, and other notifications. Light indicators were suggested to capture attention during moments of distraction or as deadlines approached. Participants also discussed expressive gestures for encouragement, including hugging, patting on the back, giving a thumbs-up, and offering a high five. One team also proposed adding a layer of physical touch by giving the robot comforting, warm “paws”, which would enhance the mental well-being of the users.

\subsubsection{Advantages and Disadvantages of a Productivity-Focused Robot}
In individual interviews, participants were asked about their preference for studying alone or in groups. Most expressed a preference for studying individually, as they felt group settings could create pressure, making them worry about bothering others or feeling behind due to varying work paces. They also found studying with friends distracting. However, many participants valued peer tutoring services at the institution for resolving doubts or discussing challenging topics, acknowledging that peers often offer additional insights and resources. Similarly, the Student Success and Well-being Coach encouraged group study to complement classroom learning and support academic growth. 
Therefore, many students saw a productivity-focused robot as an ideal substitute for study groups. P5 highlighted that a robot would not have academic concerns of its own, nor would it introduce any sense of competition or pressure, unlike studying with peers. P10 added, “Sometimes you want the social aspect (filling in the extra brain) without side conversations, so the robot would be helpful.” P17 further noted that college students often experience loneliness, and a robot could serve as a comforting companion in such times.
However, some participants raised concerns about the robot’s feasibility. They expected it to be costly, making applications on their phones a more practical alternative. Some felt that carrying a robot around the college campus might be inconvenient. One participant doubted its effectiveness, feeling uncertain about how a robot could address their productivity struggles when they themselves were not sure of the root cause.

\section{DISCUSSION}
\subsubsection{Integrating Cognitive Profiles into Inclusive Design}
The designs of all prototypes were informed by the distinct cognitive profiles of the participant groups. Research shows that individuals with ADHD are highly prone to distractions, prefer working in solo environments, and tend to favor immediate rewards, a characteristic often explained by temporal discounting \cite{scheres2021adhd, getmarlee2025}. In line with these findings, Bimo embodies a reactive design. Participants noted that ``having the robot constantly start conversations would get annoying and overwhelming, especially when the student is trying to focus and would prefer a quiet environment.'' Accordingly, Bimo minimizes unsolicited interactions while monitoring for careless mistakes and providing immediate feedback and rewards to address attentional lapses common in ADHD individuals \cite{gmehlin2016attentional}.
In contrast, studies suggest that neurotypical individuals generally function well within structured and collaborative environments \cite{mondellini2023behavioral}. Accordingly, Max is designed to promote a sense of shared progress by incorporating features such as physical gestures (e.g., hugs or high-fives) and study music, which foster emotional connection and reinforce the collaborative setting. Additionally, neurotypical users are more likely to focus on long-term benefits compared to ADHD individuals who require immediate rewards. Max offers effective study tips that reinforce this approach by promoting gradual, sustained improvement.

The mixed prototypes, Roberto and Bammy, bridge the gap by integrating reactive components, such as immediate distraction alerts, with proactive features like guidance and stress monitoring. This dual approach supports a broader spectrum of cognitive needs and underscores the benefits of involving both ADHD and neurotypical perspectives in the design process. Overall, integrating diverse cognitive perspectives into our design process promises to enhance user satisfaction in future productivity-focused SAR while laying a strong foundation for inclusive robotic systems.

\subsubsection{Essential Characteristics for a Productivity Robot}
Based on the content analysis of the group workshop, an ideal productivity-focused robot should, at minimum, possess the following characteristics:
\begin{itemize}
    \item \textbf{Small and Portable Design.} The robot should be compact and easy to move so that students can easily carry it to their preferred study location.
    \item \textbf{Friendly yet Assertive Mentor.} It should balance being a supportive companion and a motivating mentor, helping students stay on track without feeling overbearing.
    \item \textbf{Integrated Tablet.} It should be equipped with a tablet to display schedules, to-do lists, reminders and integrate with the student’s work ecosystem.
    \item \textbf{Multimodal Communication.} It should communicate with students through speech and visual displays on screens, creating a natural and interactive user experience.
    \item \textbf{Task Organization and Prioritization.} It should help users organize and prioritize tasks to improve productivity.
    \item \textbf{Stress Monitoring and Support.} It should monitor for stress and provide supportive interventions when needed.
    \item \textbf{Manual Activation for Privacy.} It should require manual turning on and off to give students full control over its operation and to respect their privacy.
    \item \textbf{No Data Storage.} It should not store any user data to ensure complete privacy and trustworthiness.
\end{itemize}
\section{ETHICAL DESIGN GUIDELINES}
\begin{itemize}
    \item \textbf{Data Privacy}. This was a central concern, as many participants raised risks around creating personal profiles from shared information. Hence, the robot’s developers must be transparent from the outset about what data is confidential. While some participants were comfortable with the robot accessing essential information to tailor productivity strategies, they preferred that the personal data collection remain optional to ensure privacy and user control. Additionally, the robot should regularly delete collected data and ensure users can access all stored information, reinforcing their sense of control over personal data.
    \item \textbf{User Autonomy and Freedom}. Another core concern shared by participants was preserving their independence and critical thinking. The robot should enhance productivity without taking away a user’s ability to think and act on their own. As P1 expressed, ``If technology takes away your thinking, it also takes your independence.'' The robot must avoid being directive and should not suggest tasks unless explicitly asked, as P20 mentioned. This would prevent the robot from subconsciously controlling the user’s day and ensure that it remains a tool that supports, rather than overrides, personal decision-making and autonomy.
    \item \textbf{Professional Boundaries}. Participants noted that, while the robot could offer general support, it should not cross the line into addressing sensitive mental health issues. The robot should be designed to recognize when a conversation goes beyond the scope of productivity coaching and refer users to qualified professionals if necessary. The Student Success and Well-Being Coach echoed these concerns, explaining that the robot cannot fully replicate the depth of human connection and understanding. It should, therefore, direct users to appropriate resources for support when needed, ensuring that it functions within ethical boundaries related to mental health and well-being.
    \item \textbf{Emotional Sensitivity and Responsiveness}. The robot should be designed to respond to users' emotional states with sensitivity and critical thinking. The robot should avoid being pushy and not overwhelm users with tasks when they are feeling stressed or struggling. The robot’s tone and interactions should always remain sensitive, supportive, and empathetic, ensuring no negative impact on the user’s emotional well-being.
    \item \textbf{Inclusivity and Cultural Sensitivity}. The robot must be inclusive and culturally sensitive, accommodating users from diverse backgrounds and experiences. It should avoid bias in its interactions and foster an environment where everyone can interact comfortably. It should also have accessibility features for people with disabilities.
    \item \textbf{User Empowerment}. The robot should focus on empowering users rather than performing tasks for them. As P21 pointed out, ``The robot shouldn’t fill the gap of the user but help them become a better version of themselves.'' The robot’s purpose should be to guide users in improving their productivity, supporting them in achieving their goals, and assisting them in managing their time, rather than doing the tasks for them and impacting their capabilities.
    \item \textbf{Complement Human Connections}. The robot should be designed to complement not replace human connections. Participants emphasized the importance of the robot reminding users not to become overly reliant on it, particularly for emotional support. It should encourage users to engage with their social circles and seek human interactions when necessary, rather than substituting these.
\end{itemize}

\section{CONCLUSIONS}

This study explored the design of a productivity-focused SAR using a participatory design approach with college students and a Student Success and Well-Being Coach. Participants responded positively to the concept. Key findings highlight students’ productivity challenges, design considerations for an effective SAR, and ethical guidelines for its development. While the study offers valuable insights, its findings are limited by the use of a participant pool from a single university. These insights will inform the development of productivity-focused social SARs in our future research.

\addtolength{\textheight}{-12cm}   







\bibliographystyle{IEEEtran}
\bibliography{sample-base}

\end{document}